\documentclass[runningheads]{llncs}
\usepackage{graphicx}
\usepackage{amsmath,amssymb} 
\usepackage{color}
\usepackage[width=122mm,left=12mm,paperwidth=146mm,height=193mm,top=12mm,paperheight=217mm]{geometry}

\usepackage{times}
\usepackage{epsfig}
\usepackage{graphicx}
\usepackage{amsmath}
\usepackage{amssymb}
\usepackage[breaklinks=true,bookmarks=false]{hyperref}
\usepackage{url}            
\usepackage{booktabs}       
\usepackage{amsfonts}       
\usepackage{nicefrac}       
\usepackage{microtype}      
\usepackage{stmaryrd}
\usepackage{dsfont}
\usepackage{algorithm}
\usepackage{algorithmic}

\usepackage{xspace}

\usepackage{wrapfig}
\usepackage{enumitem}
\usepackage{wrapfig}
\usepackage{multirow}
\usepackage{multicol}
\usepackage{subcaption}
\usepackage{hyperref}
\usepackage[utf8]{inputenc}
\usepackage{kotex}
\usepackage{xcolor}

\usepackage{amssymb}
\usepackage{pifont}
\newcommand{\cmark}{\ding{51}}%
\newcommand{\xmark}{\ding{55}}%

\begin{document}
\pagestyle{headings}
\mainmatter
\def\ECCV18SubNumber{***}  

\title{Cascaded Pyramid Network for \\ 3D Human Pose Estimation Challenge} 



\author{Sungeun Hong\textsuperscript{}, 
Wonjin Jung\textsuperscript{}, 
Ilsang Woo\textsuperscript{}, Seung Wook Kim \textsuperscript{}}
\institute{\textsuperscript{}SK T-Brain}

\maketitle
\begin{abstract}
Over the past decade, there has been a growing interest in human pose estimation.
Although much work has been done on 2D pose estimation, 3D pose estimation has still been relatively studied less.
In this paper, we propose a top-bottom based two-stage 3D estimation framework.
GloabalNet and RefineNet in our 2D pose estimation process enable us to find occluded or invisible 2D joints while 2D-to-3D  pose estimator composed of residual blocks is used to lift 2D joints to 3D joints effectively.
The proposed method achieves promising results with  mean per joint position error  at 42.39 on the validation dataset on  `3D Human Pose Estimation within the ECCV 2018 PoseTrack Challenge.'
\end{abstract}

\vspace{-0.1in}
\section{Introduction}
To tackle the challenging 3D pose estimation problem, considerable efforts have been devoted, and these can be divided into two categories.
One-stage approaches directly learn the 3D poses from monocular RGB images. 
Early investigations based on convolutional neural networks (CNN) involve a multi-task framework that jointly
trains pose regression and body part detectors.
Several subsequent approaches consider volumetric prediction and monocular model based on semantic representations \cite{zanfir2018monocular}
On the other hand, two-stage approaches first estimate 2D poses and then lift 2D poses to 3D poses. 
These approaches are motivated by the results that the influence of 2D pose information is significant in 3D pose estimation \cite{park20183d}. 
Among them, simple yet effective residual networks \cite{martinez2017simple} that directly estimate 3D poses from estimated 2D  pose results show state-of-the-art performance despite its simple architecture.

\begin{figure}
\centering
\textbf{}\includegraphics[height=5.35cm]{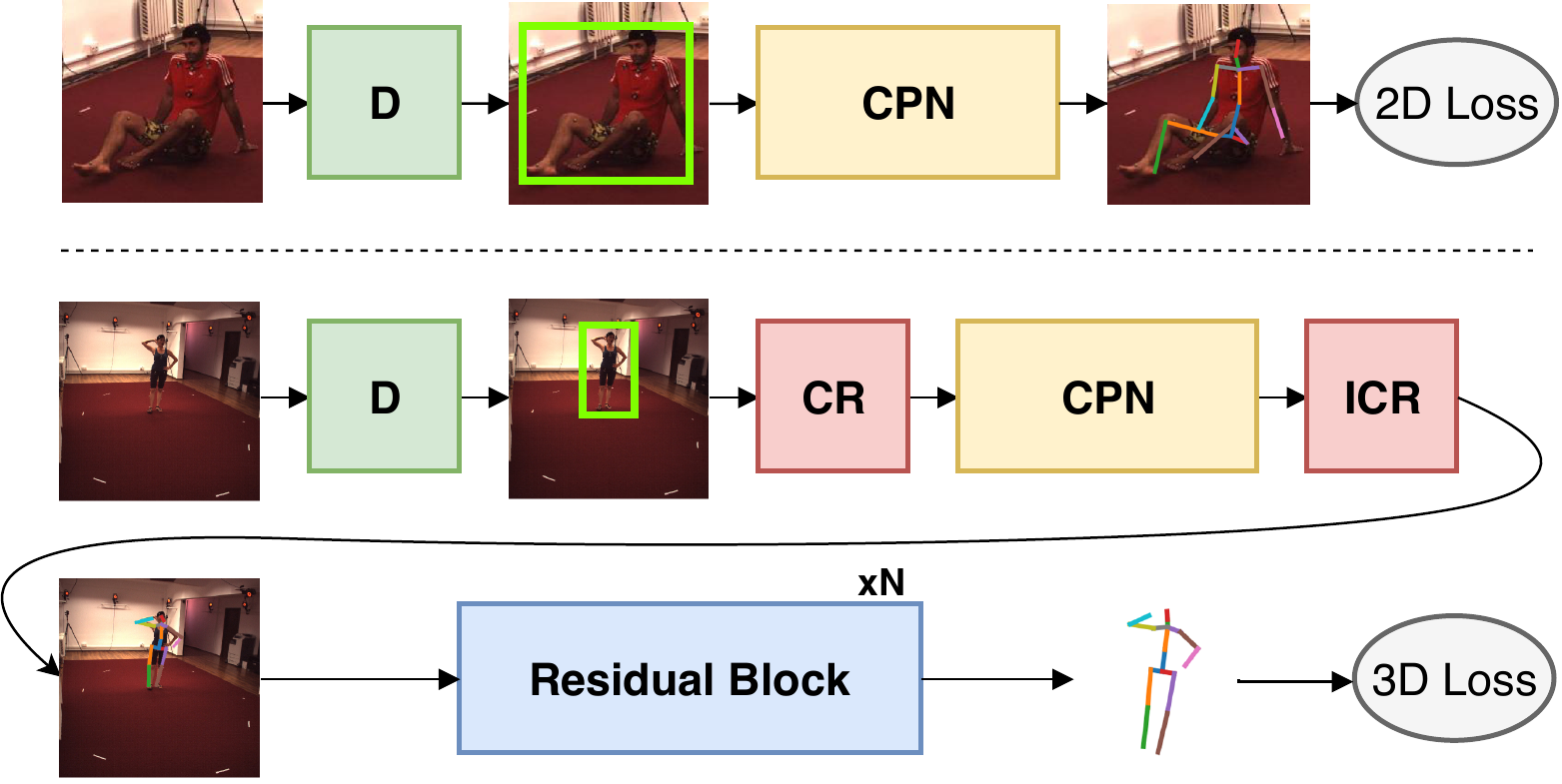}
\caption{Outline of the proposed two-step 3D pose estimation.
For 2D pose estimation, we apply a cascaded pyramid network (CPN) after detecting a person from the image.
Since the challenge dataset does not include labels for the 2D pose, a subset of Human3.6M dataset is used for 2D estimation.
We then get 2D poses of the challenge dataset using the trained 2D pose model.
To accommodate the relative size differences of the subjects between the Human3.6M and challenge dataset, we add crop and resize (CR) module before and after the CPN during the inference process.
Finally, we train the 2D-to-3D pose estimator composed of residual blocks.}
\label{fig:overview}
\vspace{-0.15in}
\end{figure}

In this paper, we propose a  top-bottom based two-stage 3D estimation framework for `3D Human Pose Estimation within the ECCV 2018 PoseTrack Challenge.'
Fig.~\ref{fig:overview} shows the overall flow of the proposed framework.
Our two-stage method achieves outstanding results with  mean per joint position error (MPJPE) at 42.39 on the validation dataset on 3D human pose estimation challenge.

\vspace{-0.1in}
\section{Proposed method}
\label{sec:ban}

\subsection{2D pose estimation}
We estimate 2D poses with a top-bottom pipeline.
Since the challenge dataset does not include labels for the 2D pose, a subset of the Human3.6M dataset \cite{ionescu2011latent,ionescu2014human3} is used to train the 2D pose estimator.
Given monocular images, we first perform human detection using Single Shot MultiBox Detector (SSD) \cite{liu2016ssd}.
We then estimate 2D poses by using cascaded pyramid networks (CPN) \cite{chen2017cascaded} which consist of  GlobalNet and RefineNet.

GlobalNet based on feature pyramid networks first localizes the keypoints in the detected bounding box. 
U-shape structure with intermediate supervision in GlobalNet helps to maintain both the spatial resolution and semantic information. 
In order to precisely estimate occluded or invisible keypoints, we apply RefineNet trained on an online hard keypoints mining loss. RefineNet transmits the information across different levels and then integrates the information of different levels.
Both GlobalNet and RefineNet generate probability heatmaps equal to the number of joints, i.e., 17 in Human3.6M.
Finally, we pick the output corresponding to the maximum probability value for each joint to estimate their positions.

Once the 2D detector is trained, we can get 2D keypoints of the challenge dataset.
As shown in Fig.~\ref{fig:overview}, the subject is tightly cropped in the Human3.6M image while  images in the challenge dataset contain a significant amount of background.
Considering the relative size differences of the subjects from the two databases, we add crop and resize (CR) module before and after the CPN in the inference process.
Concretely, the CR module generates a square based on the length of the longest side of the width or height based on the detected bounding box.
At this time, we add a little margin to the longest side to prevent the subject from being cropped too tightly.
This cropped area is then resized to 224x224 and fed into the CPN.
The process of adjusting the output of the CPN to the original scale can be processed in the reverse order of the CR. We define this as inverse crop and resize (ICR).

\subsection{2D to 3D pose estimation}
Given a 2D pose from the input image, we aim to learn a mapping function as:
\begin{equation}
  f^* = \min_f \frac{1}{N} \sum_{i=1}^{N} \mathcal{L} \left( f(\mathbf{x}_i) -  \mathbf{y}_i \right),
\end{equation}
where $\mathbf{x}_i \in \mathbb{R}^{2n}$, $\mathbf{y}_i \in \mathbb{R}^{3n}$, and $N$ is the number of the sample batch.
Following \cite{martinez2017simple}, we focus on deep neural networks based on residual blocks with batch normalization.
As a preprocessing step, we apply a standard normalization to the 2D inputs and 3D outputs by subtracting the mean and dividing by the standard deviation. 
We also zero-center both 2D and 3D poses around the hip joint.
To stabilize training, we also apply a max-norm constraint on the weights of each layer, which is efficient when coupled with batch normalization.

\vspace{-0.1in}
\section{Experiments}
\vspace{-0.05in}
\subsection{Experimental setup}
We evaluate the proposed method on the `3D Human Pose Estimation within the ECCV 2018 PoseTrack Challenge'. 
This challenge dataset consists of a training set (35,832), a validation set (19,312) and a test set (24,416).

\vspace{-0.05in}
\subsection{Quantitative results}
 We performed an ablative study to better understand the impact of each module in the proposed framework.
Table~\ref{tab:ablation} shows mean per joint position error (MPJPE) on the validation set.
The performance change with regard to the capacity of the network can be seen in Fig.~\ref{fig:capacity}.
From the experimental results, we can confirm that the performance is sufficient with one residual block with 1024 dimensions.

 \begin{minipage}{\textwidth}
  \begin{minipage}[b]{0.51\textwidth}
    \centering
    \footnotesize
  \begin{tabular}{ccccccc}
 \\
\toprule
Max-norm  & Batch-norm & Residual & Avg (w.o. CR) \\
\midrule
\xmark  & \xmark & \xmark & 59.57 (61.71) \\
\xmark  & \xmark  & \cmark & 50.05 (50.39) \\
\xmark   & \cmark& \xmark  & 51.67 (51.71) \\
\xmark   & \cmark& \cmark& 45.13 (45.85) \\
\cmark   & \xmark  & \xmark  & 122.9 (125.3) \\
\cmark  & \xmark  & \cmark & 121.6 (123.2) \\
\cmark   & \cmark & \xmark  & 50.41 (52.53) \\
\cmark   & \cmark & \cmark & 43.72 (45.41) \\
\bottomrule
\end{tabular}
\vspace{3mm}
      \captionof{table}{MPJPE on the validation set}
\label{tab:ablation}
\end{minipage}
    \begin{minipage}[b]{0.45\textwidth}
        \includegraphics[width=2.2in]{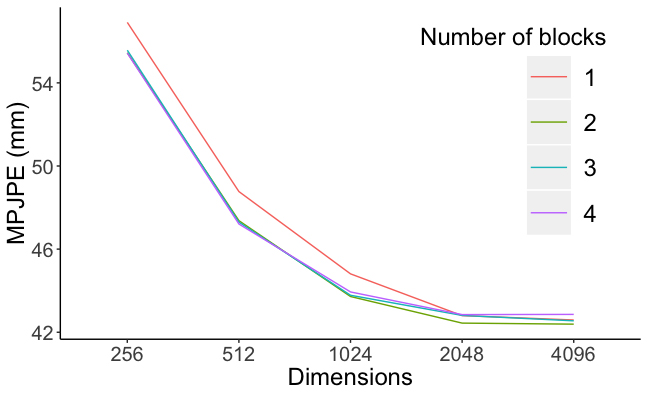}
       \captionof{figure}{Performance change as a function of the number of residual blocks and its dimensions}
     \label{fig:capacity}
  \end{minipage}
  \end{minipage}


\begin{figure}
\centering
\includegraphics[width=4.7in]{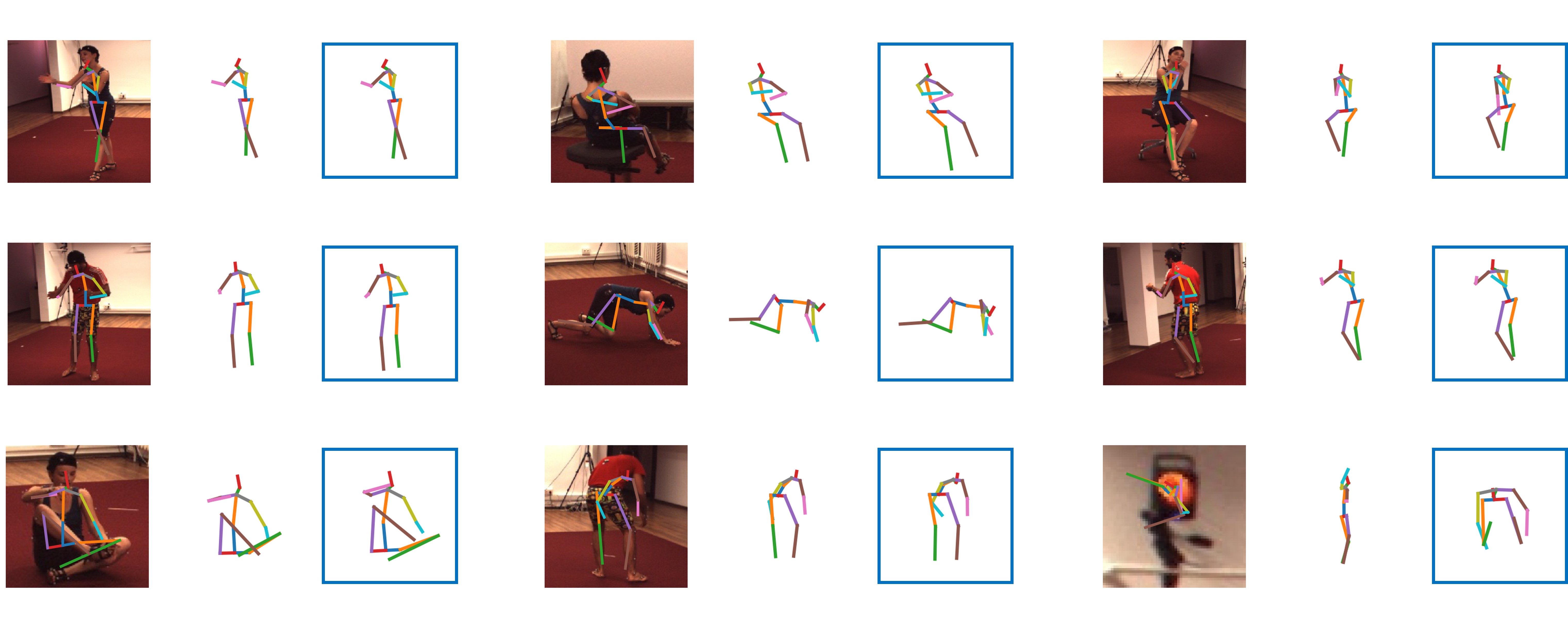}
       \captionof{figure}{Example outputs on the challenge dataset including input images with 2D pose   prediction (left), 3D pose prediction (middle), and  ground truth 3D pose (right)}
     \label{fig:qualitative}
\end{figure}
 
\subsection{Qualitative results}
We show some qualitative results on the challenge set in Fig.~\ref{fig:qualitative}. As can be seen in the figure, in most case, the proposed method shows promising results in both 2D and 3D pose estimation. 
Our top-bottom approach is detector-dependent, which affects the performance of subsequent processes, e.g., 2D pose estimation and 3D pose estimation, as can be seen in the figure.
We plan to change the proposed method to a bottom-up approach for future works.

\vspace{-.1in}
\section{Conclusions}
We propose a 3D pose estimator based on two-stage strategy composed of cascaded pyramid networks for 2D pose estimation, and residual blocks for 2D-to-3D estimation.
GloabalNet and RefinNet in cascaded pyramid networks were used to find occluded or invisible 2D joints while residual block based estimator was used to lift 2D joints to 3D joints effectively.
Our method achieves promising results with MPJPE at 42.39 on the validation dataset on 3D human pose estimation challenge.

\vspace{-.1in}
\bibliographystyle{splncs}
\bibliography{egbib}

\end{document}